# Global-Scale Self-Supervised Spatiotemporal Learning for NDVI Time-Series Reconstruction

***Ang Li*[a], *Menghui Jiang*[a], *Xiaobin Guan*[a], *Dong Chu*[b,c], *Huanfeng Shen*[a,d,e*]**

[a] *School of Resource and Environmental Sciences, Wuhan University, Wuhan 430079, China*

[b] *School of Geography and Tourism, Anhui Normal University, Wuhu 241002, China*

[c] *Key Laboratory of Earth Surface Processes and Regional Response in the Yangtze River Basin, Wuhu 241002, China*

[d] *Key Laboratory of Geographic Information System of Ministry of Education, Wuhan 430079, China*

[e] *Key Laboratory of Digital Cartography and Land Information Application of the Ministry of Natural Resources, Wuhan 430079, China*

***Corresponding author.**

*E-mail address:* shenhf@whu.edu.cn (H. Shen)

**Abstract**: Accurate and efficient reconstruction of cloud-contaminated and noise-corrupted Normalized Difference Vegetation Index (NDVI) time series remains a challenge in remote sensing. Deep learning provides a promising solution for modeling complex spatiotemporal dependencies; however, its application is often limited by the difficulty of obtaining paired clear-sky and degraded NDVI data for identical spatiotemporal locations. To address this issue, we propose GloSSR, a **Glo**bal-scale **S**elf-supervised **S**patiotemporal framework for NDVI **R**econstruction. The framework constructs supervisory signals by artificially degrading relatively clean NDVI observations with realistic cloud contamination patterns, producing self-supervised training pairs that closely mimic real-world degradation. It further introduces an end-to-end spatiotemporal learning network that jointly captures long-range temporal dependencies and short-term spatiotemporal correlation through a bidirectional Transformer with a Convolutional Long Short-Term Memory (ConvLSTM) architecture. A temporal-channel attention-based reconstruction module is incorporated to enhance informative features, while a spatiotemporal prior constraint is designed to preserve both fine-scale structures and long-term phenological trends during optimization. Extensive evaluations on moderate resolution imaging spectroradiometer (MODIS) NDVI data demonstrate the effectiveness of the proposed framework across both artificial and real-world scenarios. In artificial degraded-pixel reconstruction experiments, GloSSR achieves mean absolute errors of 0.0273, 0.0243, and 0.0279 across three representative regions, consistently outperforming the comparison methods. Time-series analyses based on real observations further demonstrate that the proposed framework can accurately characterize vegetation dynamics and capture the key phenological states. Long-term vegetation trend analysis and the transferability analysis to Advanced Very High Resolution Radiometer (AVHRR) data validate the scalability of the framework and illustrate its broad applicability for large-scale environmental monitoring.

## 1. Introduction

As a critical satellite-derived parameter, long-term normalized difference vegetation index (NDVI) time series have been extensively utilized in numerous applications, such as phenological monitoring (Fischer, 1994), ecosystem dynamics analysis (Moody and Johnson, 2001), global climate assessment (Balzter et al., 2007), and carbon cycle studies (Yu et al., 2003). However, adverse atmospheric conditions frequently introduce extensive data gaps and noise, severely degrading the temporal continuity and limiting the reliability of NDVI data in downstream analyses (Qian et al., 2025). Therefore, developing accurate and robust NDVI reconstruction methods to enhance data completeness and usability represents a crucial research imperative (Lu et al., 2025).

Over the past decades, a wide range of reconstruction approaches has been developed to address missing or noisy NDVI observations (Shen et al., 2015; Xiong et al., 2023; Zhang et al., 2024), which can be broadly categorized into three groups: one-dimensional time-series filtering methods, multi-dimensional spatiotemporal reconstruction methods, and deep learning-based methods. Owing to their simplicity and computational efficiency, time-series filtering methods have been widely adopted. These methods exploit temporal dependencies to model NDVI variations through techniques such as image compositing (Cai et al., 2025), local window filtering (Chen et al., 2004; Julien and Sobrino, 2010; Ma and Veroustraete, 2006; Sun et al., 2024), global curve fitting (Jonsson and Eklundh, 2002; Beck et al., 2006), variational solutions (Atzberger and Eilers, 2011; Liu et al., 2022; Chu et al., 2022), and frequency-domain transformations (Lu et al., 2006; Zhou et al., 2015). However, their reliance on empirical parameters and the limited performance in reconstructing large spatiotemporal gaps constrain their reliability in complex scenarios (Li et al., 2021; Ma et al., 2026). Beyond temporal filtering approaches, multi-dimensional spatiotemporal methods have been developed to jointly leverage spatial correlation (Wang et al., 2025; Padhee and Dutta, 2019; Cao et al., 2018) and the intrinsic periodicity of NDVI sequences (Chu et al., 2021; Cai et al., 2025). By integrating complementary spatial and temporal cues, these methods typically achieve higher reconstruction accuracies, particularly in heterogeneous landscapes (Yao et al., 2023). Nevertheless, their increased computational complexity often compromises efficiency and hinders scalability for large-area applications (Li et al., 2024). Therefore, balancing the accuracy and efficiency remains a critical challenge in NDVI time-series reconstruction (Li et al., 2025; Yang et al., 2022).

With the rapid development of data-driven modeling, various learning-based approaches such as XGBoost (Li et al., 2024), artificial neural networks (Das and Ghosh, 2017), and back propagation neural networks (Li et al., 2023) have been applied for NDVI reconstruction. More recently, deep learning methods have demonstrated notable advantages, owing to their strong capacity to automatically extract hierarchical spatiotemporal features and capture complex nonlinear dynamics in NDVI sequences (Shen et al., 2021; Zhao et al., 2024; Wang et al., 2025; Shu et al., 2025). However, their performance is fundamentally constrained by the lack of paired training samples, as clean and degraded NDVI sequences cannot easily be simultaneously observed at the same spatiotemporal location in real-world satellite acquisitions (Schweden et al., 2025; Zhang et al., 2026). To this end, Li et al. (2025) employed a high-precision spatiotemporal tensor completion method (ST-Tensor) to generate training labels, achieving a good balance between accuracy and efficiency. However, ST-Tensor outputs cannot fully represent authentic NDVI data and therefore impose inherent limits on the reconstruction accuracy. Moreover, the high computational cost of ST-Tensor makes it difficult to generate training samples from diverse regions, limiting model generalizability for large-scale NDVI reconstruction. Wang et al. (2025) generated training labels from patches with minimal cloud contamination, where the few remaining gaps were first interpolated. Nevertheless, such cloud-free regions are geographically scarce and restricted in certain land-cover types and climatic conditions, which limits sample diversity and model generalizability while increasing the risk of overfitting. In summary, the restricted model accuracy resulting from paired labels provided by external algorithms remains an unavoidable bottleneck, severely limiting the generalizability and scalability of NDVI reconstruction models.

Self-supervised learning (SSL) has recently emerged as a natural solution to address the lack of paired clean-degraded NDVI samples, as it constructs supervisory signals directly from intrinsic data properties without relying on externally provided labels (Wang et al., 2025). Inspired by representative SSL frameworks such as Bidirectional Encoder Representations from Transformers (Devlin et al., 2019) and Masked AutoEncoders (He et al., 2021), several studies have applied similar masking-based strategies to remote sensing (Wang et al., 2025), and have shown promising results in applications such as crop mapping and temporal prediction (Cartuyvels et al., 2023; Gao et al., 2024; Qin et al., 2025). Building on these advancements, recent efforts have explored SSL-based frameworks tailored for NDVI time series, where missing observations are artificially simulated by masking part of the clean pixels in temporal sequences. Models are then trained to reconstruct the masked values using the masked original clean observations as supervisiory signals (Liu et al., 2024; Fu et al., 2024). However, the commonly adopted simulation of missing data usually relies on a random masking strategy at a fixed ratio, which fails to reflect the actual structured, spatially coherent, and temporally persistent characteristics of cloud contamination, leading to a mismatch between the training objectives and the actual degradation patterns (Shu et al., 2025; Choudhury et al., 2025; Yang and Huang, 2026). Additionally, these works operate primarily at the pixel level and model only temporal dependencies, neglecting the spatial correlations naturally present in remote sensing images, which restricts their ability to reconstruct spatially or temporally continuous gaps that are common under the case of persistent cloud cover in remote sensing (Liu et al., 2024; Cui et al., 2026), making the spatiotemporal learning paradigm highly feasible (Tang et al., 2025). Furthermore, the existing NDVI reconstruction models are typically trained on geographically localized datasets, with few studies focusing on sample collection at large or global scales (Tang et al., 2026; Dai et al., 2025).

To address the above issues, we propose GloSSR, a **Glo**bal-scale **S**elf-supervised **S**patiotemporal learning framework for NDVI time-series **R**econstruction, designed to ensure realistic sample construction and powerful spatiotemporal feature extraction. In detail, the GloSSR framework constructs training pairs using authentic NDVI sequences sampled globally across diverse land-cover types and climatic zones. Clean and cloudy patches are identified, and clean pixels are replaced with their cloud-contaminated counterparts with added mild noise. This process generates degraded patches that realistically simulate cloud-induced degradation, providing training pairs that reflect real-world conditions. To capture both the long-range temporal dependencies and fine-scale spatiotemporal correlations in NDVI time series, we propose a spatiotemporal learning network integrating a bidirectional Transformer with a bidirectional Convolutional Long Short-Term Memory (ConvLSTM) model, complemented by a temporal-channel attention reconstruction mechanism. A spatiotemporal prior loss is incorporated to preserve the spatial details and phenological dynamics. Extensive experiments on Moderate Resolution Imaging Spectroradiometer (MODIS) data demonstrate that the GloSSR framework can achieve a superior performance on both artificially degraded and real NDVI sequences. Applications in long-term vegetation trend monitoring and the transferability to Advanced Very High Resolution Radiometer (AVHRR) data further demonstrate the practical potential of the proposed framework.

## 2. Method

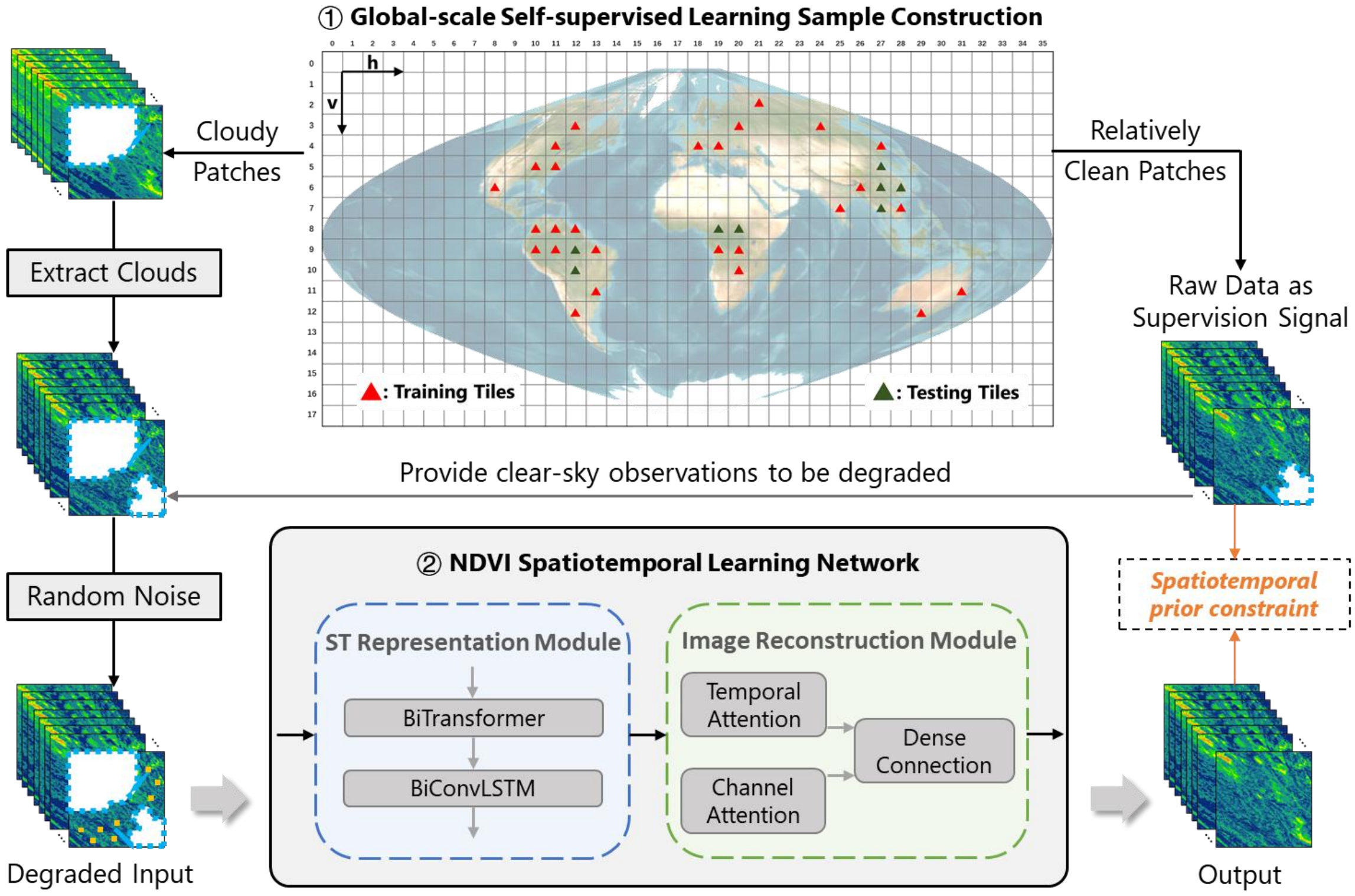

Fig. 1 Flowchart of the global-scale self-supervised spatiotemporal learning framework for NDVI time-series reconstruction.

Fig. 1 illustrates the workflow of the proposed GloSSR framework. Globally distributed self-supervised learning samples are constructed by pairing relatively clean NDVI observations with cloud-affected patches. A spatiotemporal learning network is introduced that integrates a bidirectional Transformer (BiTransformer) with a bidirectional ConvLSTM (BiConvLSTM) model to jointly capture long-range vegetation dynamics and localized spatiotemporal variations. In addition, a temporal-channel attention-based reconstruction module is developed to selectively amplify the most informative seasonal dependencies and output the reconstructed NDVI, while the spatiotemporal prior constraint regularizes the optimization process to ensure both spatial coherence and phenological plausibility.

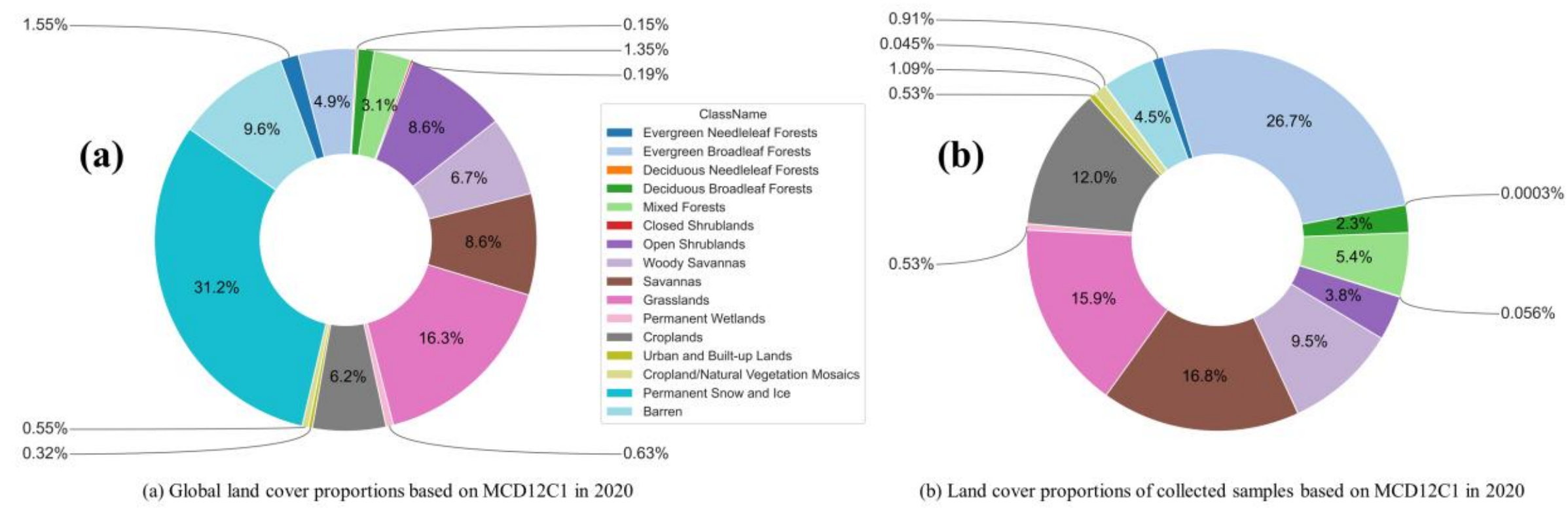

Fig. 2 Land-cover classification proportions of global (a) and collected samples (b) based on MCD12C1 in 2020.

*2.1 Global-Scale Real Data-Driven Sample Construction*

Unlike the random masking assumption, real clouds exhibit spatially continuous structures and temporally evolving patterns influenced by meteorological processes. In the GloSSR framework, to better approximate these characteristics, training pairs are constructed using globally sampled NDVI time series covering diverse biomes, climate regimes, and land-cover conditions, to ensure broad representativeness. Fig. 2(a) depicts statistics from the MCD12C1 global land-cover product for 2020. We focus on extensively sampling vegetation-related classes, as shown in Fig. 2(b), including various types of forests shrubland, savanna, cropland, etc. Clean and cloud-affected patches are identified from pixel-level quality flags using a 30% cloud-coverage threshold and then randomly paired. Cloud-free values in the clean patch are replaced with the corresponding cloud-contaminated observations, and mild perturbations are injected into the remaining clean pixels to approximate sensor or atmospheric noise. This procedure produces degraded patches that preserve realistic spatiotemporal structures and degradation patterns that are consistent with natural atmospheric interference, which allows the GloSSR to learn robust spatiotemporal features that generalize effectively at a global scale

(Miller et al., 2024).

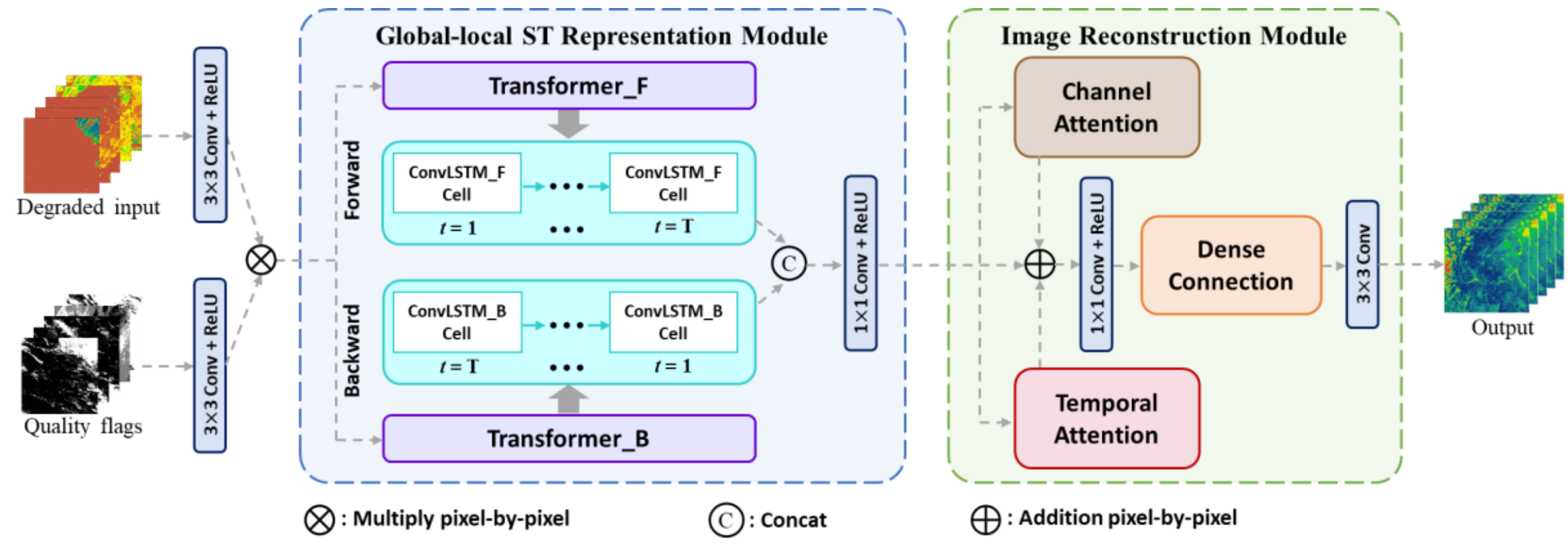

Fig. 3 NDVI spatiotemporal learning network. The suffixes F and B indicate forward and backward directions, respectively. *t* means the date number, and *T* is the temporal length of the sequence, which is 23 in this study.

*2.2 NDVI Spatiotemporal Learning Network*

The proposed NDVI spatiotemporal learning network is illustrated in Fig. 3. The degraded inputs together with the quality flags serve as network inputs, whereas the original clean patches provide the self-supervisory target. The inputs are first transformed into the feature space through two parallel 3×3 convolutions, after which the quality features modulate the NDVI features in a pixel-wise manner to explicitly encode cloud contamination. The features then pass through a global-local spatiotemporal (ST) representation module and a temporal-channel attention-based image reconstruction module to generate the seamless NDVI time series.

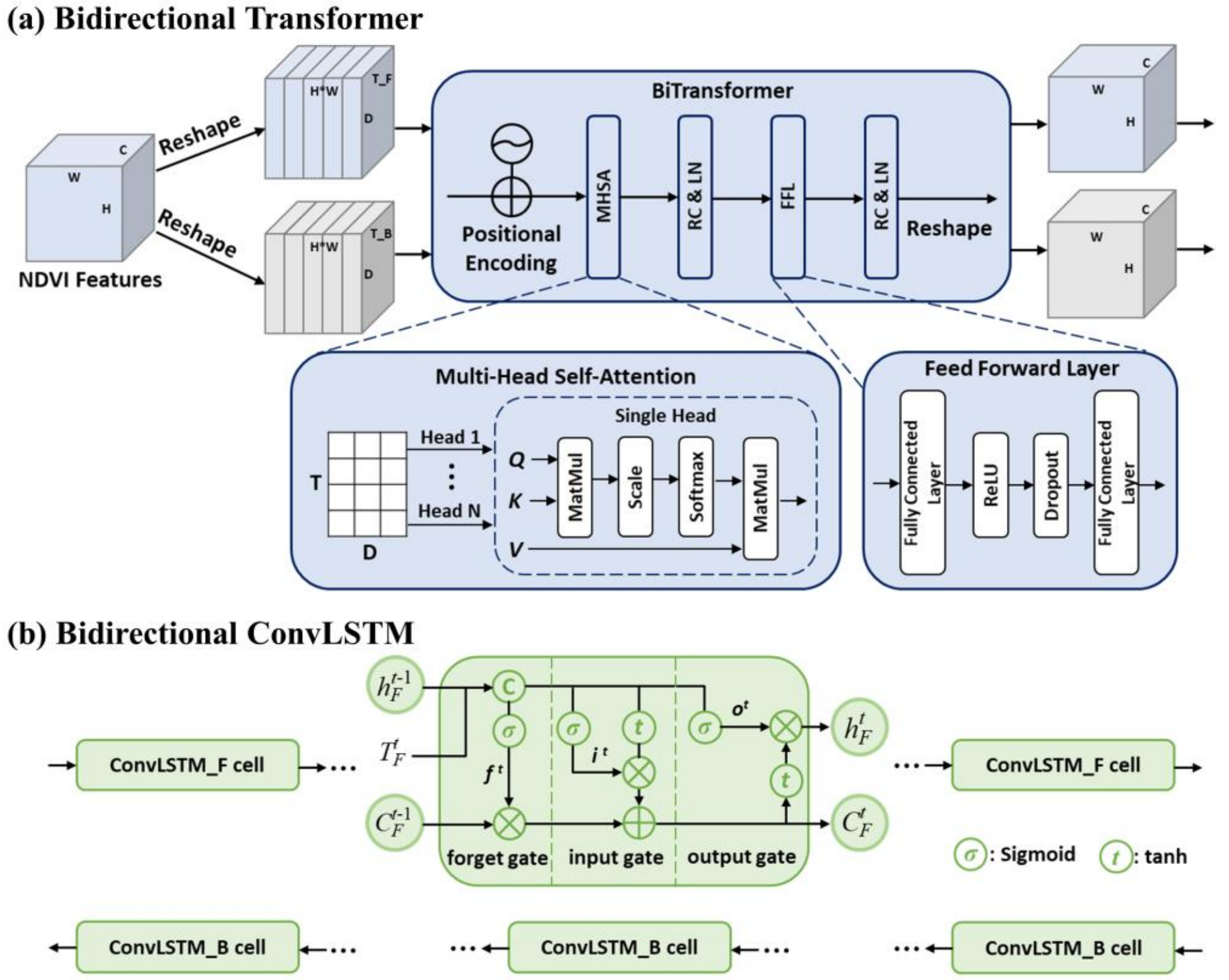

Fig. 4 The bidirectional Transformer (a) and bidirectional ConvLSTM (b) structure in the GloSSR. MHSA is the multi-head self-attention layer, RC&LN is the residual connection and layer normalization, and FFL is the feed-forward layer.

(1) Global-local Spatiotemporal Representation Module

NDVI reflects surface processes in which spatial texture and temporal dynamics are deeply intertwined. Given this point, we integrate Transformer and ConvLSTM modules to leverage their complementary strengths. The Transformer is employed along the temporal dimension to extract global long-range dependencies, while the ConvLSTM operates across the spatiotemporal dimensions to capture fine-grained, localized dynamics (Garnot and Landrieu, 2020; Vaswani et al., 2017; Sun et al., 2025). Additionally, a bidirectional design is adopted to aggregate information from both past and future contexts, enabling each time step to be modeled with complete temporal context rather than relying only on historical observations.

(a) Bidirectional Transformer: As illustrated in Fig. 4(a), the NDVI feature $N \in \mathbb{R}^{C\times W\times H}$ is reshaped into $(N_F, N_B) \in \mathbb{R}^{T\times D\times(W\times H)}$, where $H$ and $W$ represent the height and width of the NDVI image, $C$ denotes the channel dimension of the feature, $T$ is the temporal length of the sequence, and $D$ corresponds to the number of extracted NDVI image features per time step (with $C = T \times D$). $N_F$ and $N_B$ represent the features with the forward and backward time-series dimension, respectively. After reshaping, $N_F$ and $N_B$ are input into the bidirectional Transformer, which includes positional encoding (PE), a multi-head self-attention (MHSA) layer, residual connection and layer normalization (RC&LN), and a feed-forward layer (FFL). The core MHSA of the forward direction can be represented as:

$$\mathrm{MHSA}(N_{\mathrm{F}}) = [\mathrm{head}_1,\ldots,\mathrm{head}_i,\ldots,\mathrm{head}_h] \tag{1}$$

$$\mathrm{head}_i = \mathrm{Softmax}\left(\frac{W_i^Q * Q\,(W_i^K * K)^{\top}}{\sqrt{d_k}}\right) W_i^V * V \tag{2}$$

where $h$ denotes the number of heads; $i$ represents the $i$-th head; $Q$, $K$, $V \in \mathbb{R}^{D\times(W\times H)}$ denote the queries, keys, and values after linear projection of the input sequence for a given direction; $W_i^Q$, $W_i^K$, $W_i^V$ are learnable weights; $d_k = D/h$, [ ] represents concatenation; $*$ means the fully connected operation or convolution operator; and Softmax is the activation function. The complete forward Transformer output $T_{\mathrm{F}}$, after residual connection and layer normalization, is given by:

$$T_F = \mathrm{LN}\left(N_F + \mathrm{MHSA}\left(\mathrm{PE}(N_{\mathrm{F}})\right)\right) + \mathrm{FFL}\left(\mathrm{LN}\left(N_F + \mathrm{MHSA}\left(\mathrm{PE}(N_{\mathrm{F}})\right)\right)\right) \tag{3}$$

An analogous computation yields the backward representation $T_B$. The bidirectional Transformer outputs provide holistic temporal context and are passed to bidirectional ConvLSTM modules for local spatiotemporal refinement.

(b) Bidirectional ConvLSTM: In the forward ConvLSTM cell in Fig. 4(b), the input is the forward Transformer feature at time $t$, denoted $T_F^t$, the previous hidden state $h_F^{t\text{-}1}$, and the previous cell state $C_F^{t\text{-}1}$. The core mechanism of ConvLSTM is a gated memory cell that regulates information flow through input gate $i_F^t$, forget gate $f_F^t$, and output gate $o_F^t$, enabling selective retention and propagation of temporal dependencies across sequential data (Wu et al., 2025; Zhang et al., 2025; Hochreiter and Schmidhuber, 1997). The state of the ConvLSTM cell at time step $t$ is then updated as (Li et al., 2025):

$$C_F^t = f_F^t \odot C_F^{t\text{-}1} + i_F^t \odot \tanh(W * [T_F^t, h_F^{t\text{-}1}] + b) \tag{4}$$

$$h_F^t = o_F^t \odot \tanh(C_F^t) \tag{5}$$

where $W$ is the learnable convolutional parameters, $b$ is the bias, tanh is the activation function, and $\odot$ is the Hadamard product. After processing by ConvLSTM, we obtain forward global-local features $h_F$ and backward features $h_B$. These features are then concatenated and convolutional projection is applied to fuse the bidirectional information:

$$N_{TC} = \mathrm{ReLU}(W_{TC} * [h_F, h_B]) \tag{6}$$

where ReLU (rectified linear unit) represents the activation function, and $W_{TC}$ is the learnable convolutional weight. $N_{TC}$ is the output, which serves as the input to the subsequent temporal-channel attention-based reconstruction module.

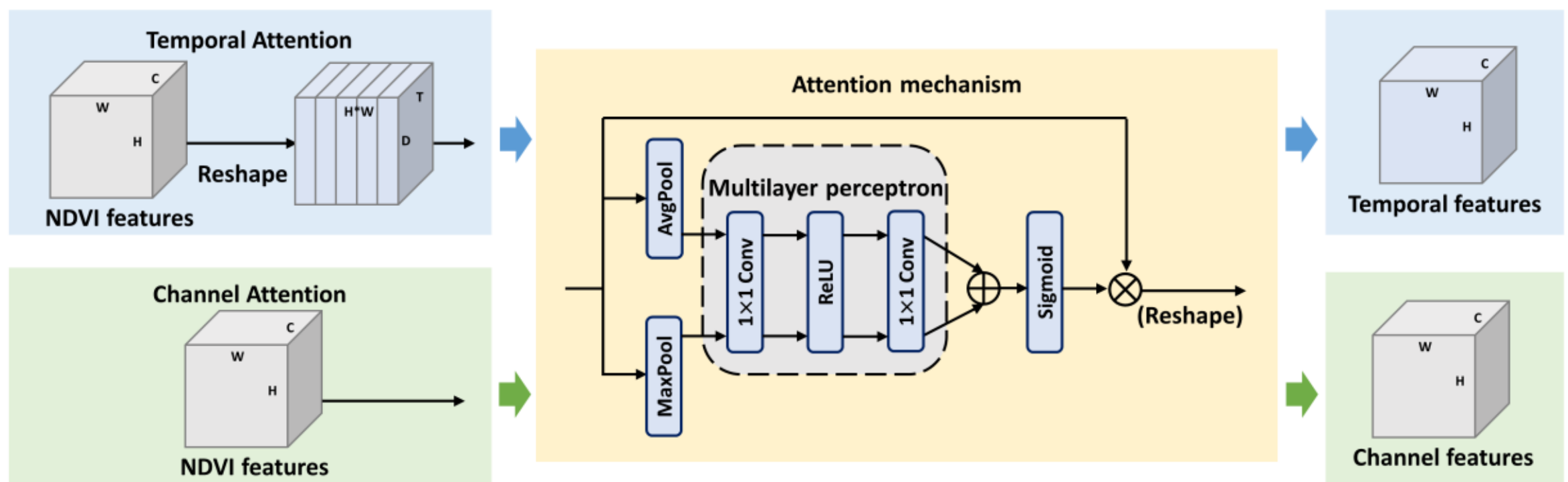

Fig. 5 Schematic of the temporal attention and channel attention mechanisms. AvgPool is the average pooling path and MaxPool is the max pooling path.

(2) Temporal-Channel Attention-Based Image Reconstruction Module

To guide the network to focus on critical phenological stages and discriminative channel patterns, we introduce a dual-attention mechanism that adaptively emphasizes informative temporal and channel responses. The input features $N_{TC}$ are processed in two complementary pathways, as shown in Fig. 5. For the temporal attention, $N_{TC} \in \mathbb{R}^{C\times W\times H}$ is reshaped into $N_R \in \mathbb{R}^{T\times D\times (W\times H)}$, explicitly structuring it as a sequence of $T$ time steps. For the channel attention, $N_{TC}$ is input into the attention mechanism, assessing the significance of each feature map generated by the preceding layers. Features from the average pooling (AvgPool) and max pooling (MaxPool) paths are linearly transformed through a multilayer perceptron (MLP), followed by a sigmoid function. The generated weights are then multiplied element-wise with the original input, enhancing the features of important time steps/channels while suppressing the influence of redundant ones (Wu et al., 2025). The temporal attention $N_{tem}$ and channel attention $N_{cha}$ are computed as follows:

$$N_{tem} = N_R \odot \mathrm{Sigmoid}\Big(\mathrm{MLP}\big(\mathrm{AvgPool}(N_R)\big) + \mathrm{MLP}\big(\mathrm{MaxPool}(N_R)\big)\Big) \tag{7}$$

$$N_{cha} = N_{TC} \odot \mathrm{Sigmoid}\Big(\mathrm{MLP}\big(\mathrm{AvgPool}(N_{TC})\big) + \mathrm{MLP}\big(\mathrm{MaxPool}(N_{TC})\big)\Big) \tag{8}$$

As illustrated in Fig. 3, the outputs of both attention mechanisms are summed with the input of the module, followed by a convolutional layer to adjust the channel dimension. The combined features are then fed into the dense connection (Dense) layer for further refinement (Huang et al., 2017). The final reconstruction output result $N_{\mathrm{rec}}$ is obtained after passing through the last convolutional layer:

$$N_{\mathrm{rec}} = W_{\mathrm{rec}} * \Big(\mathrm{Dense}\Big(\mathrm{ReLU}\big(W_{\mathrm{at}} * (N_{TC} + N_{tem} + N_{cha})\big)\Big)\Big) + b_{\mathrm{rec}} \tag{9}$$

where $W_{\mathrm{at}}$ and $W_{\mathrm{rec}}$ are the convolutional parameters, and $b_{\mathrm{rec}}$ is the bias.

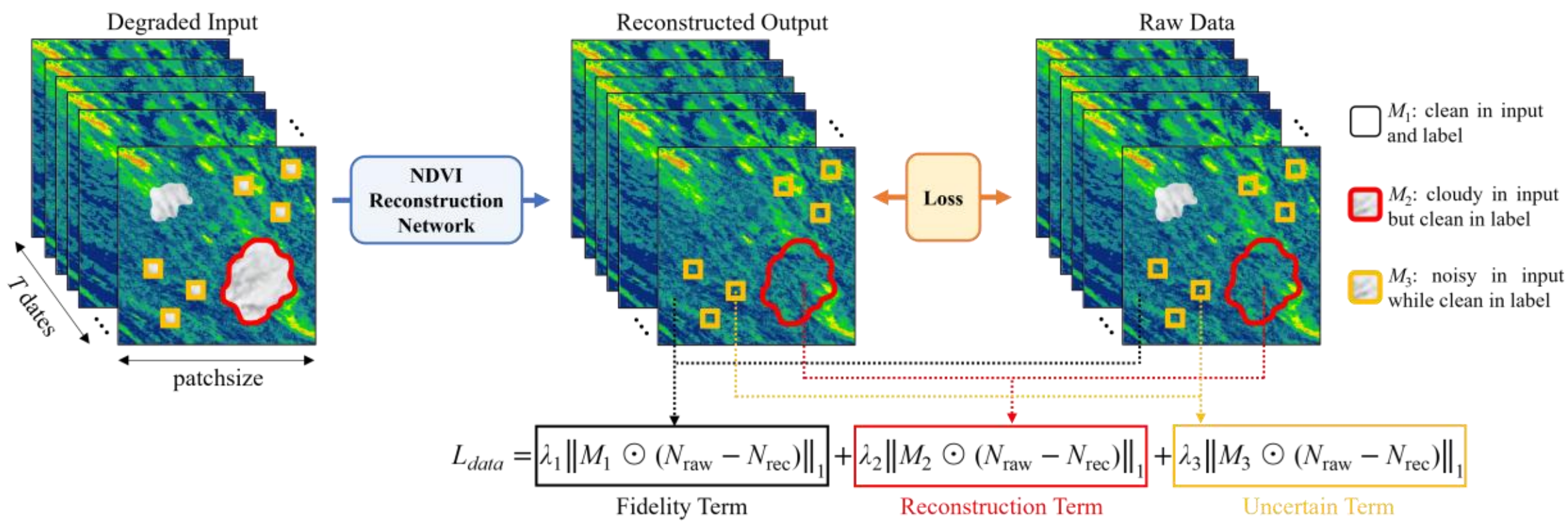

Fig. 6 The data consistency loss during network training. $M_1$ indicates clean observations shared by both the degraded input and raw data. $M_2$ corresponds to pixels that are deliberately degraded as cloud in the input but remain clean in the raw data. $M_3$ denotes pixels with added noise in the degraded input that remain clean in the raw data.

### *2.3 Loss Function*

The loss function consists of a data consistency loss and a spatiotemporal prior loss, which respectively enforce consistency with the clean NDVI time series and preserve spatiotemporal structures:

$$L = L_{cons} + L_{prior} \tag{10}$$

As shown in Fig. 6, the data consistency constraint $L_{cons}$ consists of three terms: fidelity term, reconstruction term, and uncertain term, and can be expressed as:

$$L_{cons} = \sum_{k=1}^{3} \lambda_k \left\| M_k \odot (N_{\text{raw}} - N_{\text{rec}}) \right\|_1 \tag{11}$$

where $N_{\text{raw}}$ and $N_{\text{rec}}$ denote the raw data and the reconstructed results, respectively. $M_k$ ($k$=1,2,3) represents binary masks that partition pixels into three categories. Specifically, $M_1$ indicates clean observations shared by both the degraded input and raw data, enforcing a fidelity term in Fig. 6. $M_2$ corresponds to pixels that are deliberately degraded as cloud in the input but remain clean in the raw data, defining a reconstruction term that guides the recovery from artificial degradation. $M_3$ denotes pixels with added noise in the degraded input that remain clean in the raw data, forming an uncertain term that improves the robustness to noisy observations. The coefficients $\lambda_k$ control the relative contributions of the three terms.

Furthermore, $L_{prior}$ comprises two components: a spatial gradient loss $L_{grad}$ and a temporal smoothness loss $L_{smooth}$:

$$L_{prior} = \delta L_{grad} + \varepsilon L_{smooth} \tag{12}$$

The spatial gradient loss $L_{grad}$ is designed to preserve local spatial structures and edge information (Li et al., 2025):

$$L_{grad} = \left\| M_1 \odot (\nabla_{\mathrm{p}} N_{\mathrm{raw}} - \nabla_{\mathrm{p}} N_{\mathrm{rec}}) \right\|_1 \tag{13}$$

where p = ($x$, $y$) denotes the horizontal and vertical coordinates, and $\nabla_{\mathrm{p}}$ represents the spatial gradient operation.

The temporal smoothness loss $L_{smooth}$ enforces the continuity of NDVI dynamics by penalizing high-frequency temporal fluctuations (Chu et al., 2022):

$$L_{smooth} = \frac{1}{T} \sum_{t=3}^{T} \left\| N_{\mathrm{rec}}^{t} - 2 \times N_{\mathrm{rec}}^{t\text{-}1} + N_{\mathrm{rec}}^{t\text{-}2} \right\|_1 \tag{14}$$

where $N_{\mathrm{rec}}^{t}$ represents the reconstructed result at time $t$. This term corresponds to a second-order temporal difference, encouraging smooth and physically plausible vegetation evolution while suppressing abrupt noise (Liu et al., 2024). The weighting parameters are empirically set to $\lambda_1 = 0.6$, $\lambda_2 = 2$, $\lambda_3 = 0.1$, $\delta = 0.1$, and $\varepsilon = 0.4$.

## 3. Materials and Settings

### *3.1 Experimental Data*

The MOD13A2 NDVI product provides 23 images per year with a 1-km resolution (Didan, 2021). As shown in Fig. 1, the study data were selected globally from MODIS tiles. The red triangles represent the training tiles, and the green areas represent the test tiles, with all the data spanning 2001–2024.

(1) Training Data Selection

The training samples were mainly selected from low- to mid-latitude tiles where vegetation grows vigorously and exhibits diverse seasonal dynamics. Some samples from high-latitude tiles were included to enhance representativeness. Other tiles predominantly covered by bare soil, desert, or ice/snow throughout the year were excluded, as the NDVI signatures lack the temporal variability essential for learning vegetation phenology.

(2) Test Data Selection

To facilitate a robust quantitative assessment, we constructed artificial test datasets covering three geographically distinct regions, as shown in Table I: East Asia, West Africa, and South America. These regions were chosen for their diverse vegetation types, complex climatic conditions, and prevalent cloud cover, presenting challenging scenarios for NDVI reconstruction.

**Table 1**

Overview of test dataset construction for the quantitative evaluation

| Region | Environmental characteristics | Target tile | Cloud source tile |
|---|---|---|---|
| East Asia | Diverse vegetation types, complex phenological patterns, strong seasonal variability | h27v05 | h28v06 |
| West Africa | Near equator, persistent cloud cover, humid tropical climate | h20v08 | h19v08 |
| South America | Near Amazon River plain, complex climate and terrain, severe cloud cover | h12v10 | h12v09 |

*3.2 Experimental Settings*

Training dataset construction: training data were clipped on a per-year basis along the temporal dimension, resulting in patches of size 32×32×23.

Testing dataset construction: the quantitative evaluation was conducted using artificial datasets constructed by injecting missing values from neighboring tiles into target tiles (Gerber et al., 2018; Militino et al., 2019). The specific configuration for the three test regions is summarized in Table I. In total, 10% of the clean pixels in each test dataset are randomly selected and injected with noise (Chu et al., 2022). Furthermore, for the qualitative evaluation, specific pixels and subregions from selected tiles were analyzed through detailed multi-year temporal trajectories and regional visual inspection.

Evaluation metrics and baselines: in the quantitative evaluation, the original clean observations were used as validation data to calculate the evaluation metrics, including the correlation coefficient (CC), root-mean-square error (RMSE), and mean absolute error (MAE). The comparison methods were the SG filter (Chen et al., 2004), HANTS filter (Zhou et al., 2015), and ST-Tensor (Chu et al., 2021). The SG filter is a widely used method for NDVI time-series reconstruction, the HANTS filter is a commonly used frequency-domain algorithm, and ST-Tensor is an advanced multi-dimensional variational method.

## 4. Evaluation and Results

### *4.1 Results on Artificial Data*

Table 2 presents the average quantitative evaluation results for each method across the years 2001–2024 in the different scenario, including cloudy pixels (cloud removal), noise pixels (denoising), and the overall comparison (overall) with the raw clean observations. Bold text indicates the best result, and underlined text shows the second-best result. Higher CC, lower RMSE, and lower MAE indicate better quantitative evaluation results. The quantitative results across the three representative regions show that GloSSR achieves the best performance under all metrics and scenarios. In the cloud removal tasks, ST-Tensor generally ranks second, while HANTS attains the second-best CC in West Africa. The performance gap between the GloSSR and the other methods is particularly pronounced in West Africa and South America, where persistent cloud cover exists. For denoising, all the methods demonstrate a better performance than for cloud removal, with SG and ST-Tensor showing a comparable capability in handling random noise. The quantitative evaluation results demonstrate that GloSSR can obtain a stable and reliable performance across diverse geographic regions and degradation conditions.

**Table 2**

Mean quantitative evaluation results for the reconstructed pixels and validation data

| Test area | Methods | Cloud removal | | | Denoising | | | Overall | | |
|---|---|---|---|---|---|---|---|---|---|---|
| | | CC↑ | RMSE↓ | MAE↓ | CC↑ | RMSE↓ | MAE↓ | CC↑ | RMSE↓ | MAE↓ |
| East Asia | HANTS | 0.8987 | 0.0765 | 0.0539 | 0.9482 | 0.0530 | 0.0357 | 0.9468 | 0.0535 | 0.0344 |
| | SG | 0.9027 | 0.0748 | 0.0524 | 0.9718 | 0.0399 | 0.0290 | 0.9656 | 0.0436 | 0.0282 |
| | ST-Tensor | 0.9577 | 0.0440 | 0.0321 | 0.9720 | 0.0374 | 0.0283 | 0.9835 | 0.0285 | 0.0208 |
| | GloSSR | **0.9624** | **0.0397** | **0.0273** | **0.9850** | **0.0278** | **0.0207** | **0.9907** | **0.0207** | **0.0124** |
| West Africa | HANTS | 0.6702 | 0.0645 | 0.0468 | 0.9424 | 0.0384 | 0.0280 | 0.9400 | 0.0387 | 0.0252 |
| | SG | 0.6552 | 0.0658 | 0.0472 | 0.9574 | 0.0310 | 0.0233 | 0.9423 | 0.0341 | 0.0203 |
| | ST-Tensor | 0.4889 | 0.0516 | 0.0355 | 0.9579 | 0.0314 | 0.0233 | 0.9476 | 0.0293 | 0.0192 |
| | GloSSR | **0.8738** | **0.0366** | **0.0242** | **0.9762** | **0.0236** | **0.0177** | **0.9839** | **0.0174** | **0.0098** |
| South America | HANTS | 0.8551 | 0.0676 | 0.0448 | 0.9610 | 0.0436 | 0.0298 | 0.9330 | 0.0512 | 0.0308 |
| | SG | 0.8465 | 0.0701 | 0.0459 | 0.9785 | 0.0338 | 0.0249 | 0.9390 | 0.0476 | 0.0277 |
| | ST-Tensor | 0.9194 | 0.0481 | 0.0327 | 0.9770 | 0.0344 | 0.0252 | 0.9703 | 0.0336 | 0.0228 |
| | GloSSR | **0.9370** | **0.0430** | **0.0279** | **0.9850** | **0.0279** | **0.0203** | **0.9810** | **0.0266** | **0.0150** |

*Note. Cloud removal means the scenario where the raw clean observations are degraded as cloudy pixels. Denoising means that the raw clean observations have noise added. Overall means that all the raw clean observations are compared before and after reconstruction.

To visually demonstrate the ability of each method, the reconstruction results from the three test regions are shown in Fig. 7, Fig. 8, and Fig. 9, respectively. The CC between each method and the clean pixels of the raw image is also provided. In Fig. 7, most pixels of the raw image are flagged as "marginal" (uncertain, potentially noisy). The major crops were generally in the growth stage at this time, rather than at their seasonal peak. It can be observed that HANTS and SG exhibit noticeable overestimation of low values within the blue-lined subregions, where most pixels are flagged as "good". ST-Tensor preserves the low values, to some extent, while GloSSR demonstrates a closer visual agreement with the raw image. In Fig. 8, vegetation was in a vigorous growth stage at this time, with the forest canopy maintaining a consistently high NDVI. While underestimation can be observed for the comparison methods in the red-lined subregions, GloSSR reconstructs image that is closer to the expected summer vegetation peaks. ST-Tensor exhibits low values in the lower-left portion of the red rectangle. In Fig. 9, the land cover is a heterogeneous mix of tropical rainforest, river valleys, and croplands. HANTS, SG and ST-Tensor methods slightly underestimate the high NDVI values in the river valley area within the red square subregion, while GloSSR produces results that are shows consistent with the raw image. Moreover, within the blue-lined area, GloSSR avoids overestimation issues in agricultural fields.

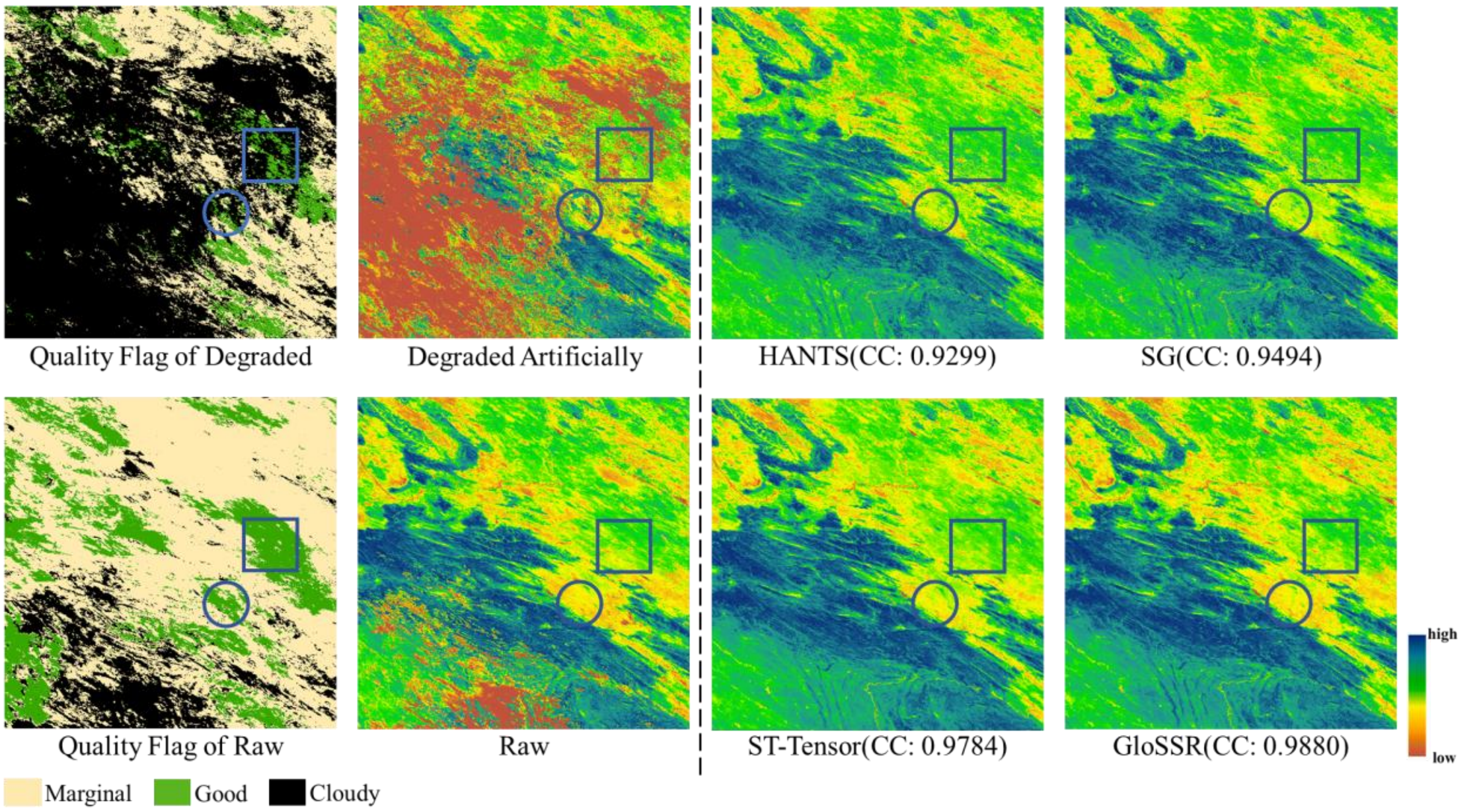

Fig. 7 The visual results for the artificial data for East Asia on May 24, 2012. The CC between each method and the clean pixels of the raw image is privided.

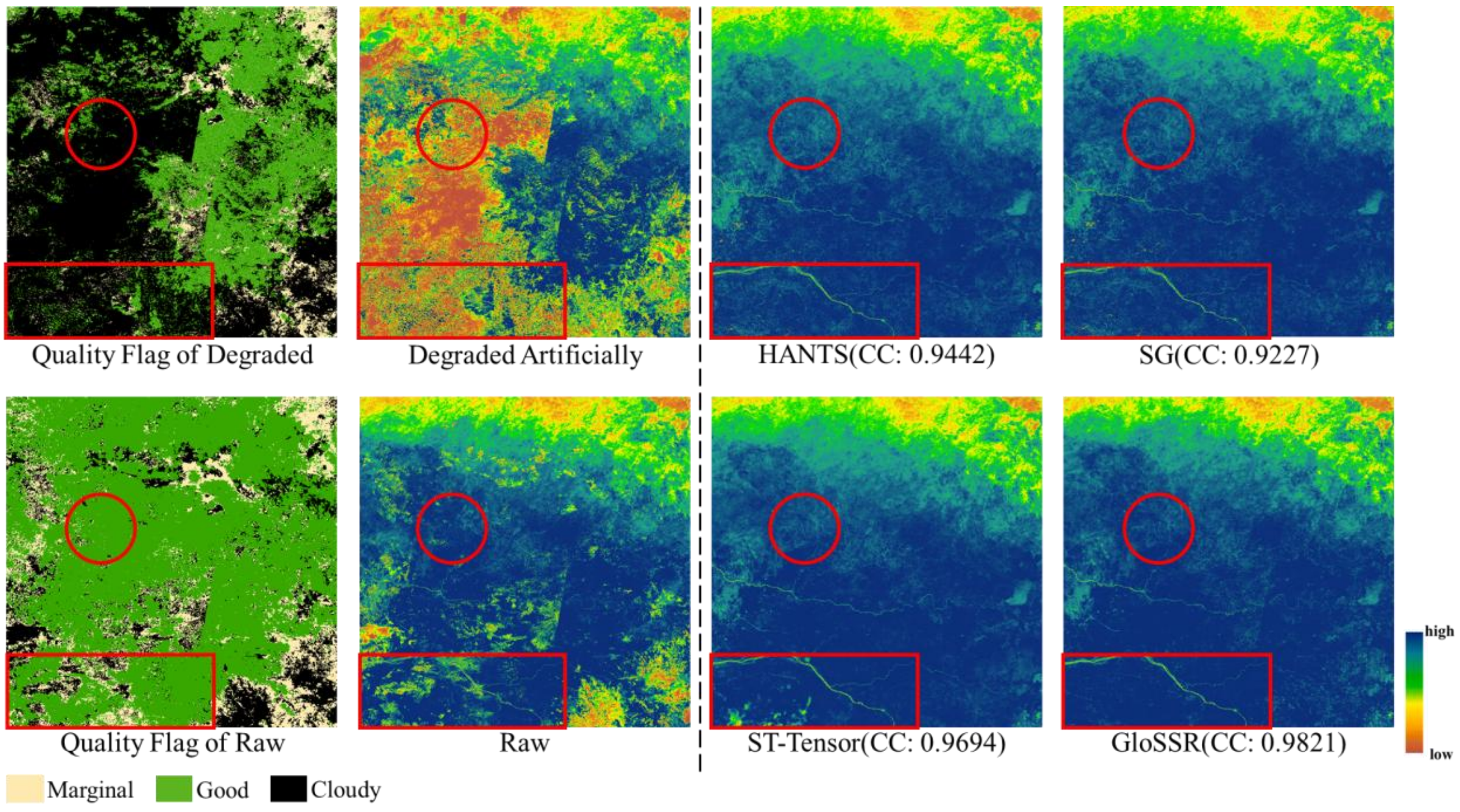

Fig. 8 The visual results for the artificial data for West Africa on June 10, 2001.

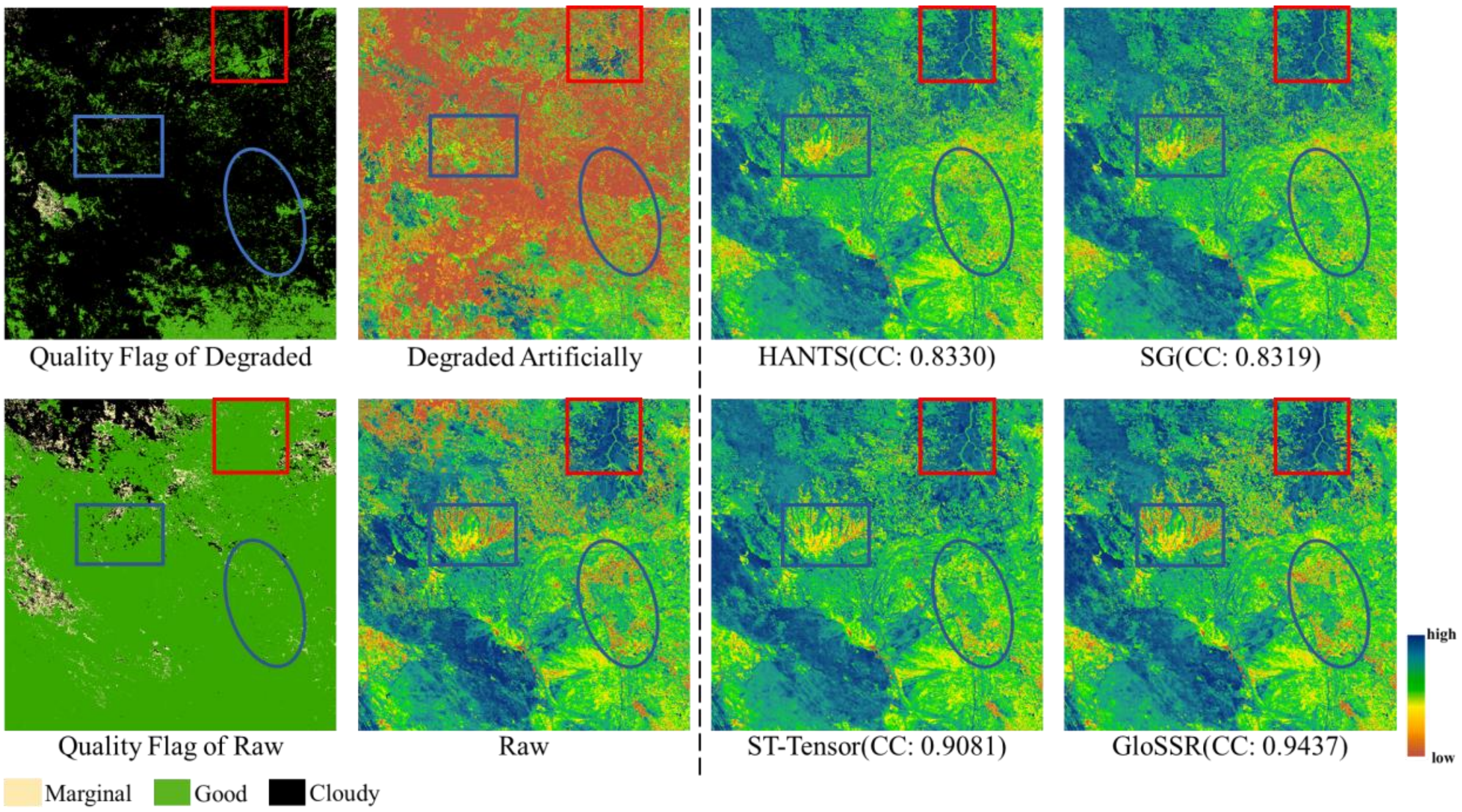

Fig. 9 The visual results for the artificial data for South America on March 6, 2017.

### *4.2 Temporal Consistency Analysis on Real Observations*

As shown in Fig. 10, representative pixels from different vegetation types are selected to display their long-term time-series variations from 2001 to 2024. The GloSSR demonstrates advantages in two aspects, the first being the reconstruction of consecutive missing data. For example, during 2008, 2012 in the savanna pixel (Fig. 10(a)), 2024 in deciduous broadleaf forest pixel (Fig. 10(c)), and 2015, 2020 in the mixed forest pixel (Fig. 10(d)), the presence of continuous missing data causes noticeable underestimation in HANTS and SG methods. ST-Tensor can extract interannual information from other years to solve the problem. However, under extremely cloudy and rainy conditions, the rainy season leads to data gaps during the same months each year, making interannual periodic information difficult to capture. In such cases, ST-Tensor has difficulty reconstructing long continuous missing segments, as shown for mid-2017 in Fig. 10(b). The GloSSR framework can effectively restore NDVI values affected by persistent cloud cover because its training samples are constructed from a large number of consecutive cloud-prone patches.

The other advantage is that GloSSR can accurately capture the phenological characteristics, which is important for vegetation dynamics and response monitoring (Wang et al., 2025; Zhang and Jin, 2021). For example, GloSSR retains the natural low-value periods associated with early growth or senescence. In mid-2003, for the cropland pixel (Fig. 10(f)), when the vegetation transitions from a decline to a growth phase and reaches a trough, HANTS and SG exhibit abnormal overestimation, whereas ST-Tensor and GloSSR successfully retain the local low values. Furthermore, GloSSR performs well in reconstructing the complex multi-peak seasonal structures. During the harvesting stage in 2002, 2013, and 2022 in Fig. 10(f), the NDVI values drop to a low level, but both the SG and HANTS exhibit overestimation to varying degrees. For both the planting and harvesting phases, GloSSR restores the NDVI values to be closer to the ground truth. Fig. 11 further presents some time-series curves for specific years,where the proposed GloSSR exhibits a robust performance in filling continuous gaps (Figs. 11(a)–(e)) and preserving critical phenological details (Fig. 11(f)).

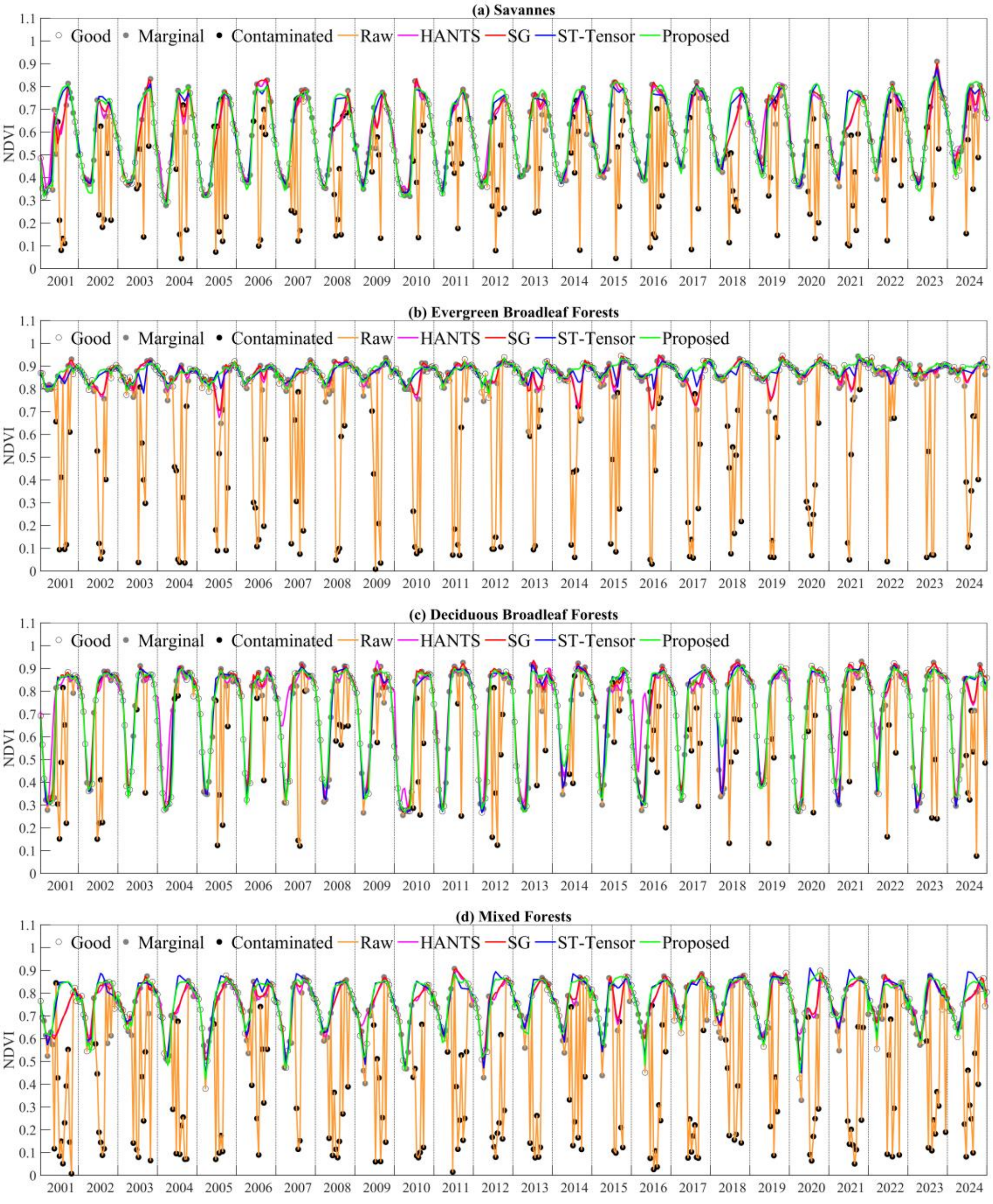

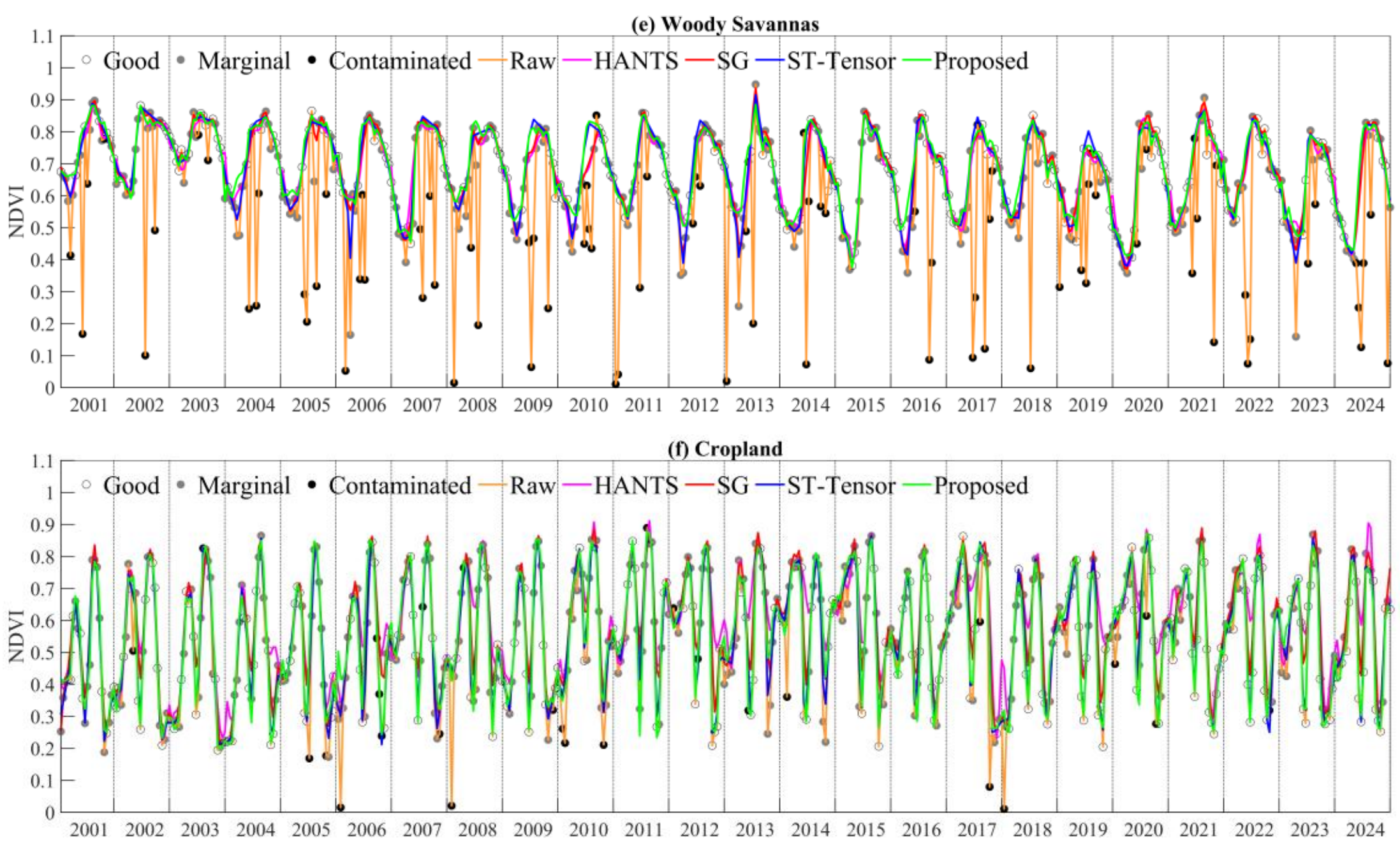

Fig. 10 Long-term NDVI change curves for typical vegetation types.

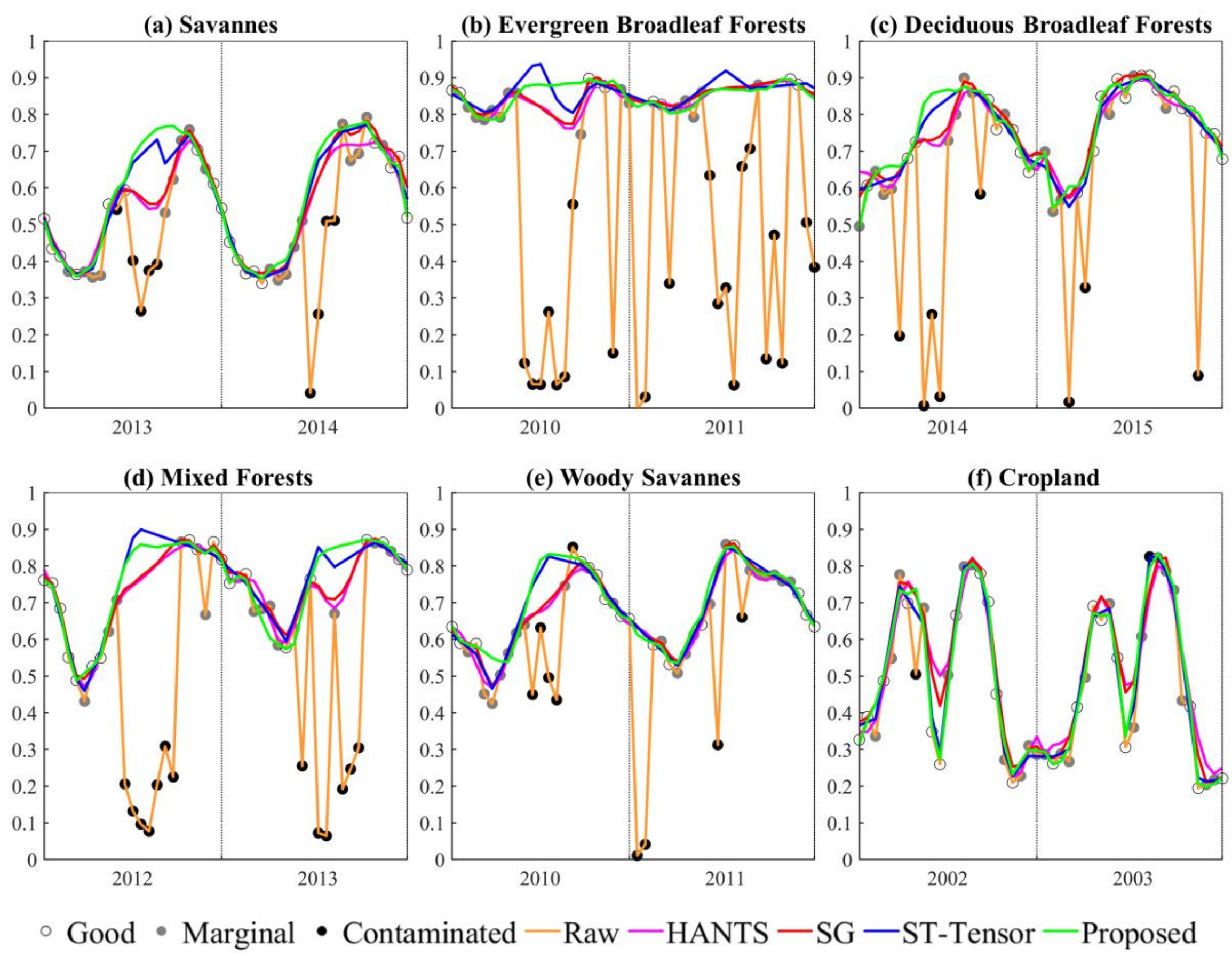

Fig. 11 NDVI change curves in specific years for several typical vegetation types.

### *4.3 Visual Spatial Pattern Assessment on Real Observations*

Test regions with severe cloud pollution are selected to show the visual results. The results for the East Asia sub-region on the 241st day of 2004 (August 28) are presented, during which NDVI values approach or reach the annual maximum, but continuous cloud cover occurs. Fig. 12 displays the original image with a missing data rate of 85.58% and the reconstructed results from each method, along with the mean pixel values $\mu$. The raw image on the 257th day of 2004 (September 13, the next observation) with less cloud is used for the visual comparison. It can be seen that each method reconstructs the cloud-covered pixels, with GloSSR more effectively restoring the NDVI to high values ($\mu$ = 0.6471), which are closer to the next observation ($\mu$ = 0.6688), followed by ST-Tensor ($\mu$ = 0.6347). Additionally, GloSSR can accurately simulate the NDVI valleys between harvest and the subsequent growing seasons. A sub-region of tile h27v05 on day 177 in 2012 (June 25th) is shown in Fig. 13, where the northern part contains cropland while the southern part features lush natural vegetation. The northern crops were in a harvesting-phase at this time, with naturally low NDVI values. The quality flags indicate that most areas were cloud-covered or marginal (uncertain, potentially noisy), with only sparse valid observations. The image acquired on the 177th day of 2011 under relatively clear conditions is also displayed. GloSSR obtains lower NDVI values for the cropland region, better matching the harvesting-phase values for the same date in 2011 ($\mu$ = 0.6172), while the other methods tend to overestimate the valley values, which corresponds to the time-series curves for the cropland in Fig. 10(f) and Fig. 11(f). The pixel average for the GloSSR ($\mu$ = 0.6588) is lower than that for the SG ($\mu$ = 0.6752) and ST-Tensor ($\mu$ = 0.6661) methods. HANTS produces the lowest average value ($\mu$ = 0.6553), primarily due to the underestimation of NDVI for the lush vegetation in the southern cloud-covered area.

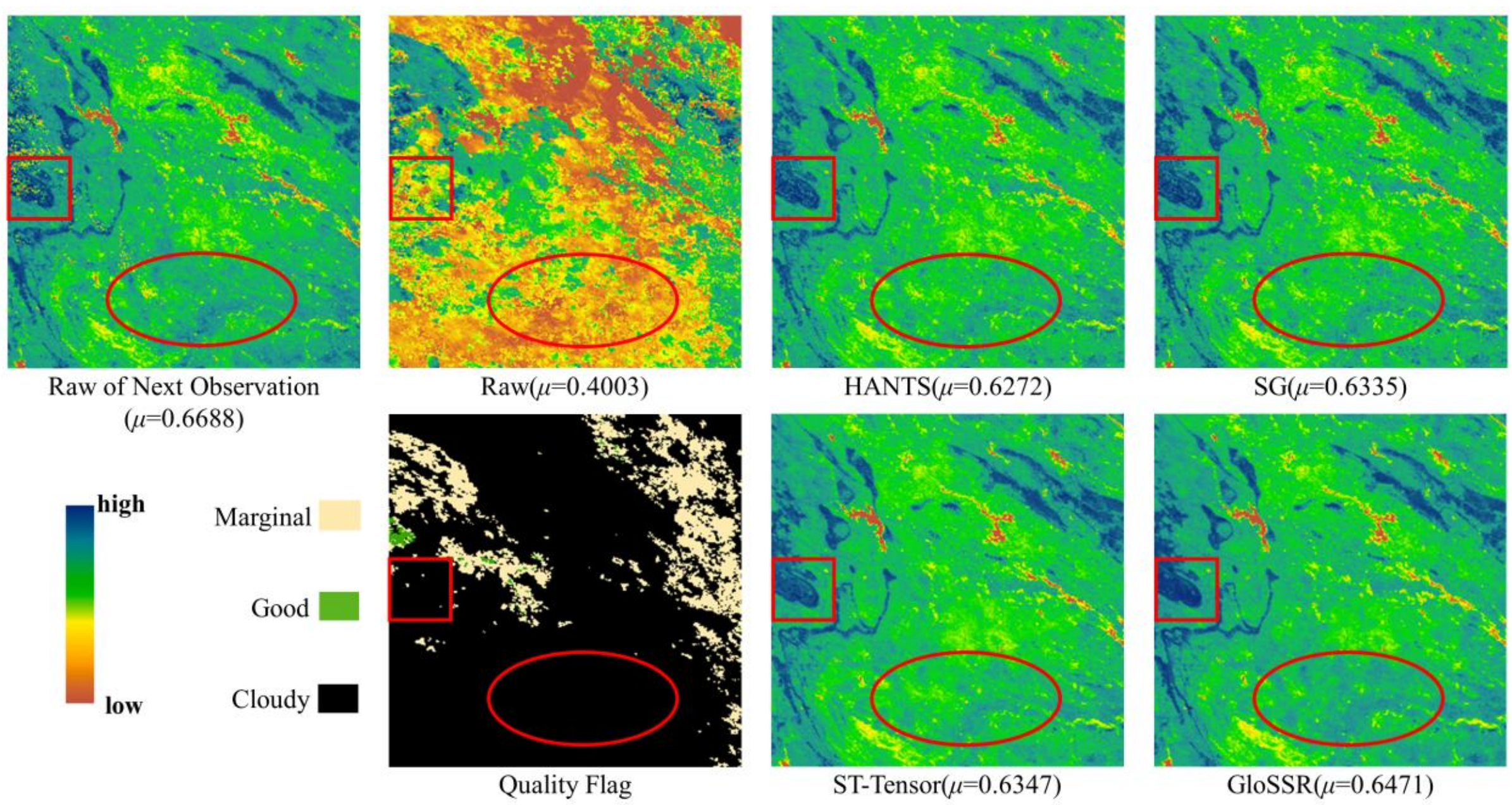

Fig. 12 Visual results for the cloudy and rainy sub-region on August 28, 2004.

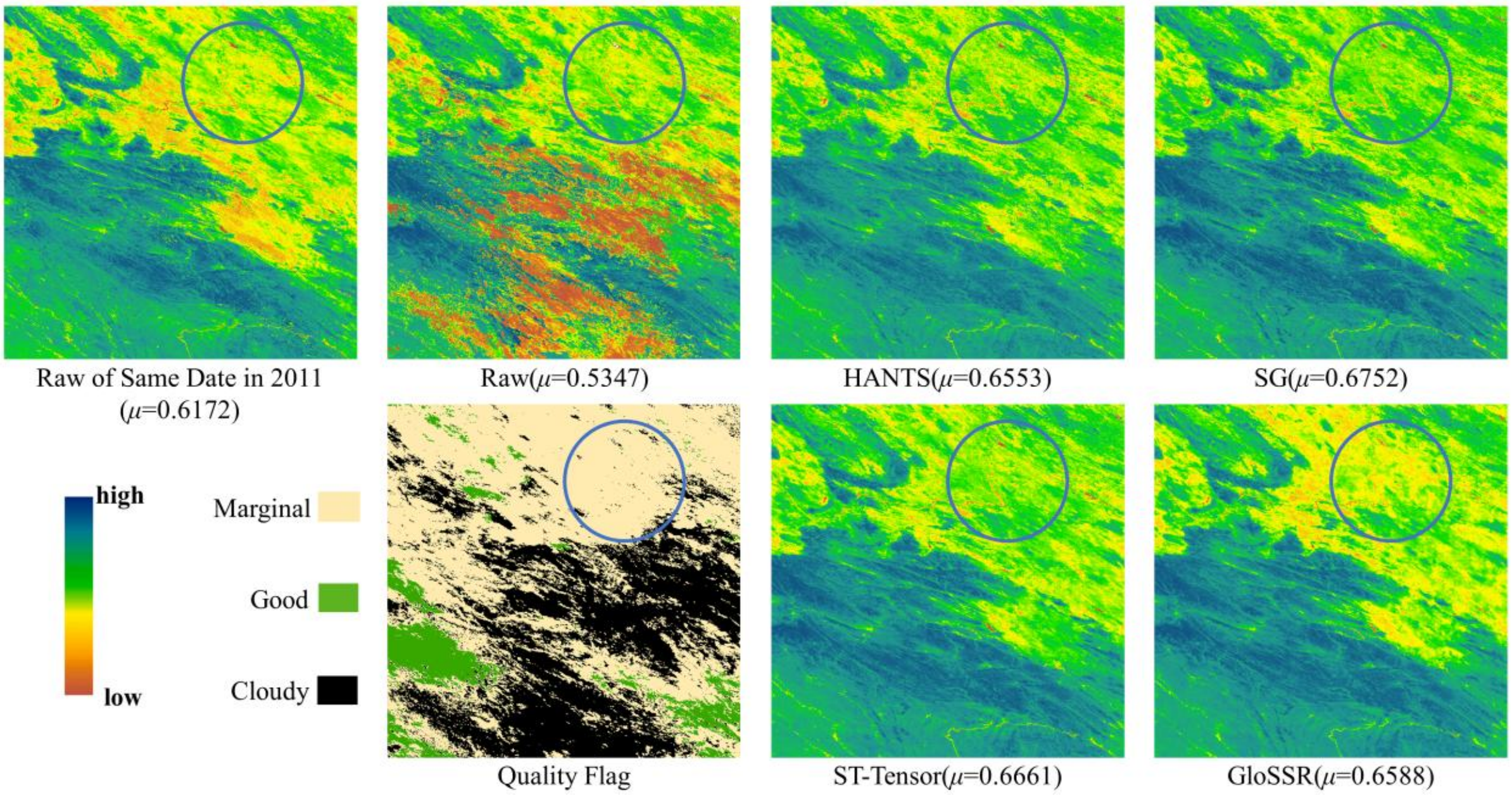

Fig. 13 Visual results for part of tile h27v05 on June 25, 2012.

## 5. Discussion

In this section, we discuss GloSSR from multiple perspectives, including the ablation experiments, operational efficiency analysis, vegetation trend analysis, and extension to AVHRR sensor data.

### *5.1 Ablation Experiments and Operational Efficiency Analysis*

To validate the effectiveness of each component in the network structure and loss function, ablation experiments were conducted using the artificial data for the West Africa test area. Table 3 lists the ablation experiment results for the cloud removal and denoising tasks. The networks correspond to versions with the Transformer, temporal attention, channel attention, bidirectional structure, and each term of the loss removed, respectively. The bold font represents the optimal values for each metric, while the underlined values represent the second-best results. Specifically, excluding the Transformer results in reduced CC and increased RMSE/MAE, especially in denoising, indicating the effectiveness of modeling long-range temporal dependencies. Compared with the channel attention mechanism, the temporal attention mechanism causes a larger performance drop, which suggests that accurately capturing the temporal correlations is more critical than cross-channel feature recalibration. The bidirectional structure plays an important role in both tasks. Removing it causes a consistent performance degradation, indicating that exploiting both past and future observations is essential for reconstructing vegetation dynamics. From the loss function perspective, the reconstruction term has the most significant impact on cloud removal. Removing the fidelity term or uncertainty term causes minor performance variations, while removing the spatial gradient loss negatively impacts the cloud removal. Notably, omitting the temporal smoothness loss yields better quantitative metrics, especially in denoising. However, as shown in Fig. 14, removing the temporal smoothness constraint ($\varepsilon = 0$) leads to a noticeably jagged result. As the value of $\varepsilon$ increases, the reconstructed time-series curves become smoother, but also weaken the temporal details. In real-world vegetation dynamics, NDVI evolution is generally gradual and phenologically continuous rather than abruptly oscillatory. Although increasing $\varepsilon$ reduces the quantitative accuracy, the decrease remains within an acceptable range. To balance the quantitative accuracy and realistic temporal dynamics, the temporal smoothness term is kept, and $\varepsilon$ = 0.4 is selected as it preserves key temporal patterns while avoiding spurious oscillations. The full model consistently achieves an excellent performance across all metrics, demonstrating the effectiveness of the integrated design.

**Table 3**

Ablation experiment results for different conditions

| Network | Cloud removal | | | Denoising | | |
|---|---|---|---|---|---|---|
| | CC ↑ | RMSE ↓ | MAE ↓ | CC ↑ | RMSE ↓ | MAE ↓ |
| w/o Transformer | 0.8606 | 0.0380 | 0.0255 | 0.9609 | 0.0314 | 0.0254 |
| w/o temporal attention | 0.8582 | 0.0379 | 0.0268 | 0.9574 | 0.0301 | 0.0241 |
| w/o channel attention | 0.8567 | 0.0370 | 0.0244 | 0.9648 | 0.0297 | 0.0238 |
| w/o bidirectional structure | 0.8510 | 0.0373 | 0.0252 | 0.9585 | 0.0359 | 0.0285 |
| w/o fidelity loss term | 0.8713 | 0.0369 | 0.0244 | 0.9691 | 0.0269 | 0.0196 |
| w/o reconstruction loss term | 0.7611 | 0.0548 | 0.0411 | 0.9710 | 0.0255 | 0.0193 |
| w/o uncertain loss term | 0.8718 | 0.0371 | 0.0245 | 0.9686 | 0.0281 | 0.0204 |
| w/o spatial gradient loss | 0.8657 | 0.0391 | 0.0266 | 0.9745 | 0.0243 | 0.0177 |
| w/o temporal smoothness loss | **0.8747** | **0.0358** | **0.0231** | **0.9872** | **0.0169** | **0.0127** |
| Full GloSSR framework | <u>0.8738</u> | <u>0.0366</u> | <u>0.0242</u> | <u>0.9763</u> | <u>0.0234</u> | <u>0.0176</u> |

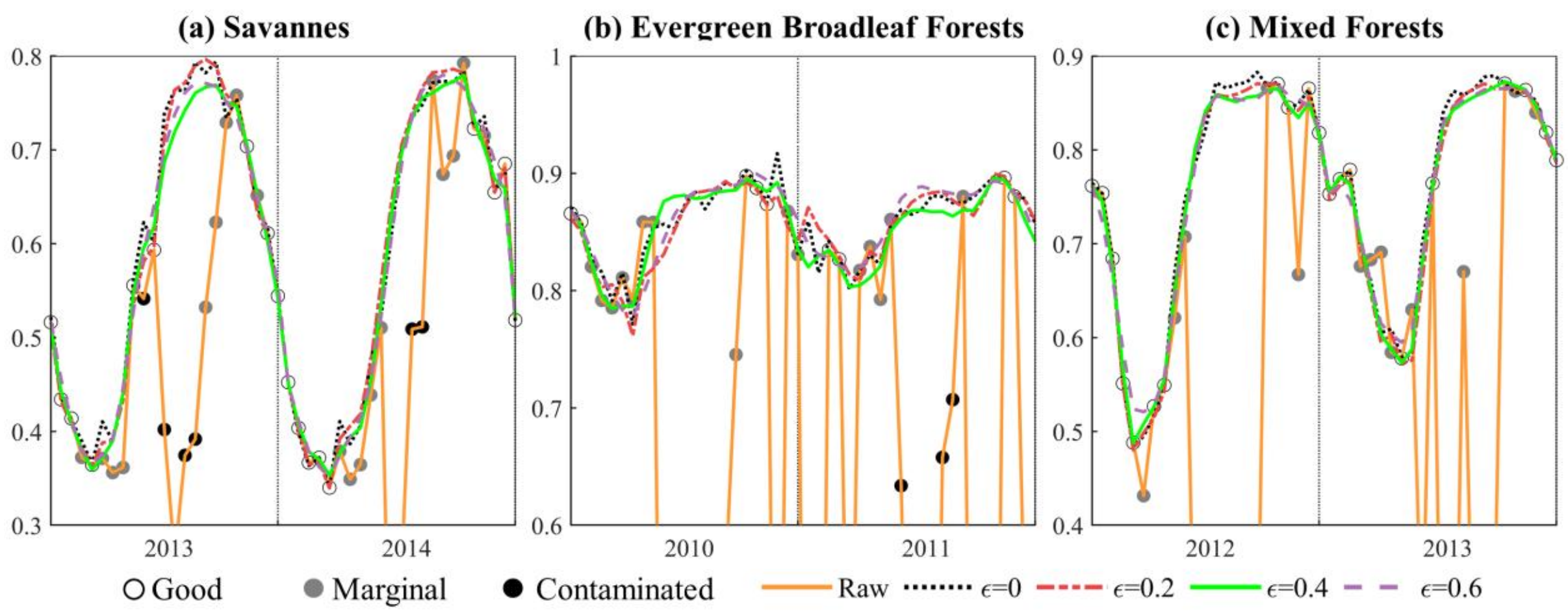

Fig. 14 Comparison of reconstructed NDVI time series using different smoothness loss weights $\varepsilon$.

We comprehensively compared the runtimes of the different methods under different CPU

core counts, with the proposed method supporting GPU-based parallel computation. The test data consisted of 24 years of time-series data, with a size of 800×800×529. Table 4 shows the running times of HANTS, SG, ST-Tensor, and GloSSR under various CPU core numbers (“c” in Table 4), as well as GloSSR under GPU conditions. The results demonstrate that SG achieves the fastest processing speed, followed by HANTS. The proposed GloSSR has a lower inference efficiency than HANTS but still maintains a competitive performance, completing the reconstruction of the 24-year time-series NDVI in around 11 minutes. Notably, GPU parallel acceleration further reduces the inference time to 126 seconds, demonstrating its strong potential for long-term time series reconstruction tasks.

**Table 4**

Running times of the different methods under different calculation conditions

| Method | HANTS | | | SG | | | ST-Tensor | | | GloSSR (proposed) | | | |
|---|---|---|---|---|---|---|---|---|---|---|---|---|---|
| Configuration | 8c | 16c | 32c | 8c | 16c | 32c | 8c | 16c | 32c | 8c | 16c | 32c | GPU |
| Time(s) | 381 | 256 | 185 | 113 | 108 | 101 | 11470 | 9996 | 9863 | 658 | 476 | 374 | 126 |

*Note. “c” means the CPU core number.

*5.2 Analysis of Vegetation Change Trends*

We reconstructed the 2001–2024 NDVI time-series for tile h27v06 based on GloSSR, and conducted a vegetation trend analysis (D’Ercole et al., 2024; Zhang and Jin, 2021) with Theil-Sen median trend analysis (Cai et al., 2024) and the Mann-Kendall trend test (Zhang and Wu, 2020). Table 5 summarizes the area percentage of the various trends under different Theil-Sen median slope ($S$) and Mann-Kendall standardized test statistic ($Z$) values. $S$ reflects the long-term monotonic trend, whereas $Z$ measures the statistical significance of that trend. The results indicate that the majority of the area (80.85%) exhibits an extremely significant increasing trend ($S > 0$, $Z > 2.58$). Non-significant changes, including both increases and decreases, account for 9.91% of the total area, while moderately and significantly increasing trends collectively represent 4.98%. Fig. 15(a) classifies the trends into nine categories based on Table 5. It can be observed that, after reconstruction, the cloud-prone mountainous region (red ellipse) transitions from a non-significant increase to a highly significant increase trend, while the rapid urban expansion region (red square) shifts from a non-significant increase to a non-significant decrease. Fig. 15(b) illustrates the spatial distribution of the Thiel-Sen median slopes. In the upper left, which is the cloud-prone mountainous region (red square), GloSSR provides a higher positive $S$, showing an increasing trend that better matches the ecological characteristics of the region (Wei et al., 2024). In the square region of the lower-right corner affected by rapid urban expansion in the Red River delta, the value of $S$ derived from the reconstructed results is markedly lower, reflecting a weaker increasing or decreasing trend in vegetation growth, which is more consistent with urbanization. The comparison between the raw and reconstructed trend patterns demonstrates that GloSSR produces trend results that better align with actual surface processes.

**Table 5**

Statistics of the vegetation change trends from 2001 to 2024

| *Slope(S)* | $Z$ | NDVI trend | Area ratio |
|---|---|---|---|
| $S$<0 | $Z$>2.58 | Extremely significant decrease | 2.97% |
| | 1.96<$Z$≤2.58 | Significant decrease | 0.82% |
| | 1.65<$Z$≤1.96 | Mildly significant decrease | 0.49% |
| | $Z$≤1.65 | Non-significant decrease | 3.71% |
| $S$=0 | $Z$ | No change | 0 |
| $S$>0 | $Z$≤1.65 | Non-significant increase | 6.20% |
| | 1.65<$Z$≤1.96 | Mildly significant increase | 1.54% |
| | 1.96<$Z$≤2.58 | Significant increase | 3.44% |

|  | $Z$>2.58 | Extremely significant increase | 80.85% |
|---|---|---|---|

*Note. $S$ is the Theil-Sen median slope, and $Z$ is the Mann-Kendall standardized test statistic value.

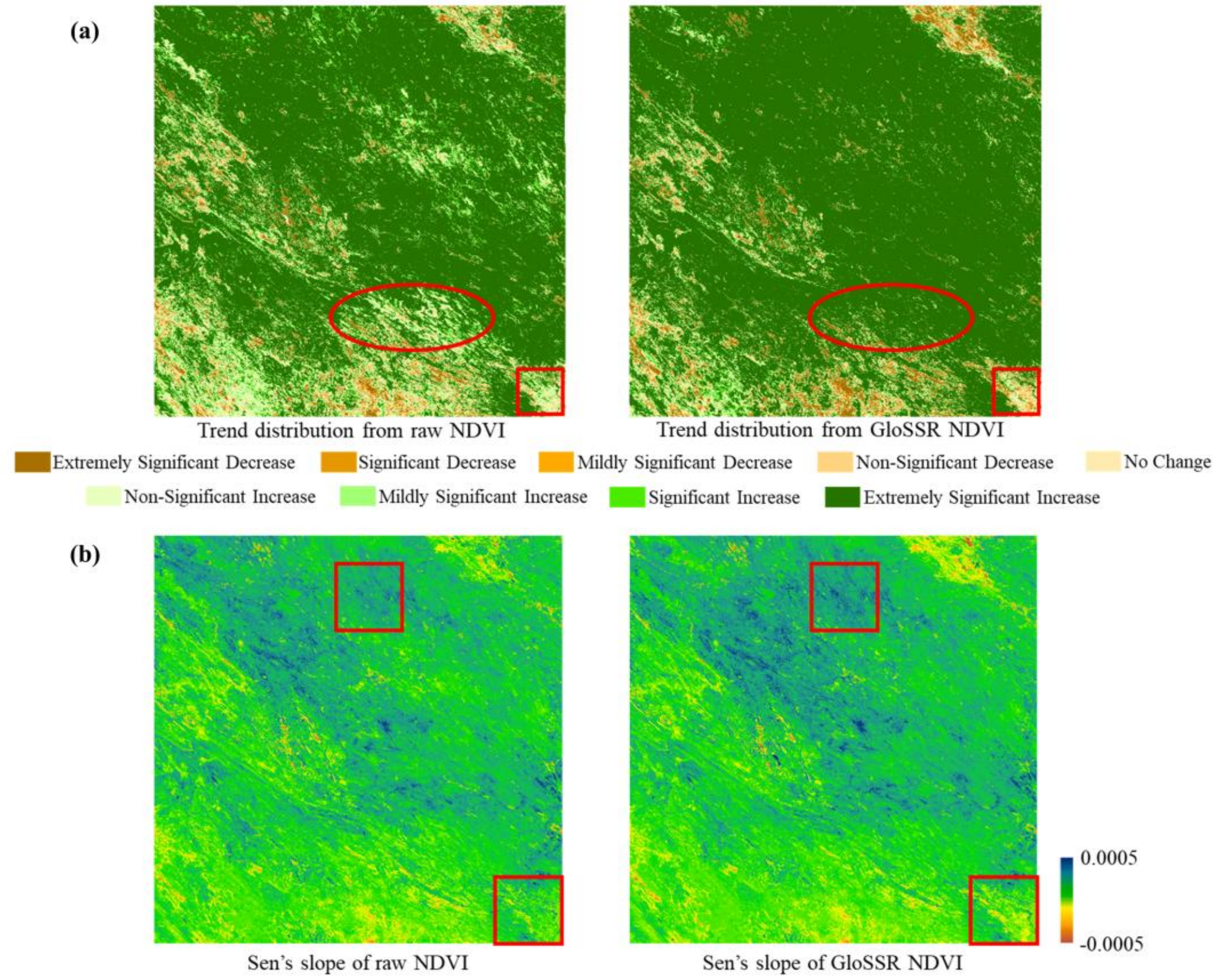

Fig. 15 Mann-Kendall trend distribution (a) and Theil-Sen slope spatial distribution (b) before and after reconstruction.

*5.3 Sensor Transferability of GloSSR*

To evaluate the generalization capability and cross-sensor robustness of the proposed GloSSR framework, we applied the model trained exclusively on MODIS data to AVHRR observations, without any fine-tuning. Global Inventory Modeling and Mapping Studies-3rd Generation (GIMMS-3G) NDVI data have a spatial resolution of approximately 0.0833° (~8 km) and are composited at a semi-monthly interval. The artificial test data for the East Asia region from 2001 to 2010 were used for the quantitative evaluation. In Table 6, it can be seen that the GloSSR achieves substantially higher CC and significantly lower RMSE and MAE values. In comparison with the second-best ST-Tensor, the CC increases from 0.6719 to 0.7586 for cloud removal, while the RMSE and MAE are reduced by approximately 25%. Similar trends can be observed for the denoising task, suggesting the effectiveness of the GloSSR in handling gaps and noise that differ from the training domain. The strong transferability of GloSSR can be attributed to the fact that the training samples are constructed from real NDVI observations, whose intrinsic vegetation evolution patterns remain consistent across different sensors. This transferability experiment demonstrates the potential of GloSSR for reconstruction tasks using data from different sensors.

**Table 6**

Transferability analysis on GIMMS-3G data

| Method | Cloud removal | | | Denoising | | |
|---|---|---|---|---|---|---|
| | CC↑ | RMSE↓ | MAE↓ | CC↑ | RMSE↓ | MAE↓ |
| HANTS | 0.5438 | 0.2534 | 0.2084 | 0.6635 | 0.1602 | 0.1229 |
| SG | 0.5565 | 0.2589 | 0.2118 | 0.6690 | 0.1535 | 0.1181 |
| ST-Tensor | 0.6719 | 0.1734 | 0.1457 | 0.6815 | 0.1434 | 0.1106 |
| GloSSR | **0.7586** | **0.1298** | **0.1077** | **0.8003** | **0.1169** | **0.0931** |

## 6. Conclusion

In this paper, we have proposed a global-scale self-supervised spatiotemporal learning framework for NDVI time-series reconstruction. The GloSSR framework leverages globally distributed NDVI data for authentic sample construction, guiding the model to learn mappings that better reflect real-world degradation scenarios. The framework introduces a spatiotemporal learning network that integrates a spatiotemporal representation module combining a bidirectional Transformer and bidirectional ConvLSTM for capturing both global-local NDVI features, along with a tailored image reconstruction module based on temporal-channel attention. A spatiotemporal prior constraint is designed to guide the optimization effectively, ensuring both spatial coherence and reasonable phenology. Extensive experiments illustrate that GloSSR can accurately reconstruct cloud-obscured pixels and suppress noise, consistently outperforming the existing methods while considerable showing good operational efficiency. A temporal analysis and the visual spatial patterns observed on real observations show that GloSSR can maintain temporal trajectory coherence and preserve fine spatial textures. The application in vegetation trend analysis and data transferability further demonstrate the practical potential of GloSSR. Although the proposed framework demonstrates strong effectiveness, there remains room for further improvement. The setting of the loss function weights may not generalize optimally to different datasets, and the reconstructed time series may sometimes be slightly less smooth, due to inherent irregularities in the raw data. In the future, adaptive learning of the weights represents a promising direction. The framework will also focus on generating large-scale remote sensing data products for precision agriculture and environmental monitoring applications.

## CRediT authorship contribution statement

**Ang Li:** Methodology, Investigation, Formal analysis, Data curation, Writing – original draft, Writing – review & editing. **Menghui Jiang:** Conceptualization, Supervision, Methodology, Funding acquisition, Writing – review & editing. **Xiaobin Guan:** Supervision, Validation, Writing – review & editing. **Dong Chu:** Conceptualization, Validation. **Huanfeng Shen:** Conceptualization, Validation, Supervision, Funding acquisition, Writing – review & editing.

## Declaration of competing interest

The authors declare no competing interests.

## Data availability

All data used in this study are publicly available: MODIS MOD13A2 data (https://

earthexplorer.usgs.gov/). Global Inventory Modeling and Mapping Studies-3rd Generation (GIMMS-3G) NDVI: (https://www.earthdata.nasa.gov/).

## Acknowledgement

The numerical calculations in this paper were performed on the supercomputing system at the Supercomputing Center of Wuhan University. This research was supported in part by the National Science and Technology Major Project of China [Grant number 2026ZD1216300] and the National Natural Science Foundation of China [Grant numbers 42571435].